\documentclass[times, 10pt,twocolumn]{article} 
\usepackage{latex8}
\usepackage{times}
\usepackage{epsfig}
\usepackage{graphicx}
\usepackage{amsmath}
\usepackage{amssymb}
\usepackage{subcaption}

\begin{document}

\title{Multi-Task Image-Based Dietary Assessment \\for Food Recognition and Portion Size Estimation}

\author{Jiangpeng He$^*$ \quad Zeman Shao$^*$ \quad Janine Wright$^{\dagger}$ \\ Deborah Kerr$^{\dagger}$ \quad Carol Boushey$^{\ddagger}$ \quad Fengqing Zhu$^*$ \\
$^{\star}${School of Electrical and Computer Engineering, Purdue University, West Lafayette, Indiana USA}\\
$^{\dagger}$ {School of Public Health, Curtin University, Perth, Western Australia, Australia}\\
$^{\ddagger}$ {Cancer Epidemiology Program, University of Hawaii Cancer Center, Honolulu, Hawaii, United States}
}


\maketitle
\thispagestyle{empty}

\begin{abstract}
Deep learning based methods have achieved impressive results in many applications for image-based diet assessment such as food classification and food portion size estimation. However, existing methods only focus on one task at a time, making it difficult to apply in real life when multiple tasks need to be processed together. In this work, we propose an end-to-end multi-task framework that can achieve both food classification and food portion size estimation. We introduce a food image dataset collected from a nutrition study where the groundtruth food portion is provided by registered dietitians. The multi-task learning uses L2-norm based soft parameter sharing to train the classification and regression tasks simultaneously. We also propose the use of cross-domain feature adaptation together with normalization to further improve the performance of food portion size estimation. Our results outperforms the baseline methods for both classification accuracy and mean absolute error for portion estimation, which shows great potential for advancing the field of image-based dietary assessment.
\end{abstract}

\Section{Introduction}
\label{introduction} 
Dietary assessment is the process of determining what someone eats and how much energy is consumed during the course of a day. It provides valuable insights for mounting intervention programs for prevention of many chronic diseases. Modern deep learning techniques have achieved great success in image-based dietary assessment for food classification~\cite{IBM,yanai2015food, deepfood-liu2016,foodnet-pandey2017} and food energy estimation~\cite{aizawa_2013,fang_2015,murphy_2015,fang2019end,icip2018}. However, existing methods only focus on one task at one time, which makes it challenging to integrate into a complete system for fast and streamlined process. In this work, we focus on designing an end-to-end multi-task framework that can identify food types and estimate their portion sizes from single image. 

Image classification is one of the most common tasks in computer vision. In image-based dietary assessment, it is important to monitor and record what kind of food people eat for disease prevention. However, estimating an object's portion size is a challenging task. An object's portion size is defined as the numeric value that is directly related to the spatial quantity of the object in world coordinates. Examples may include an object's volume and weight, as $weight \propto volume$ ($weight = volume \times density$). In food portion size estimation, we want to estimate food energy ($food\  energy \propto food\  volume$, as $food\  energy = food \  volume \times unit \ volume  \ energy $) from an input image since energy intake is an important indicator for healthy eating.

Multi-task learning aims to solve more than one tasks simultaneously, which is typically done with either hard or soft parameter sharing of hidden layers. Based on experimental results shown in Section~\ref{expresults}, hard parameter sharing is not a feasible solution for our application since it is difficult for the two tasks to share one common feature space. In this work, we introduce soft parameter sharing where each task has its own feature space and the lower layer of the two models are regularized. Our goal is to investigate the connection between the two tasks and our experimental results show that the performance of both food classification and food portion size estimation can be improved by regularizing the lower layers using L2 norm. In addition, due to the difficulty of directly mapping an RGB image to a numeric portion size, we apply cross-domain feature adaptation that concatenates the feature vectors extracted from the classification network with the feature vectors extracted from the regression network.  The feature vectors from the classification task can provide prior knowledge to better inform the portion size estimation given the food category is known. To adapt the features extracted from different domains for joint regression, we extensively studied the use of normalization techniques~\cite{ba2016layer, ioffe2015batch} in Section~\ref{CDFA}.

Success of modern deep learning based methods also rely on the availability of data. The lack of good datasets have resulted limited progress end-to-end image-based dietary assessment system. Currently, there is no available food image dataset that includes both food category and corresponding portion size since it is difficult to obtain accurate food energy from the crowd based annotation on RGB images, unless these numeric values are recorded during image collection. To address this problem, we introduce an eating occasion datasets containing both food category and food portion size provided by registered dietitians. We describe the collection of this dataset in Section ~\ref{datasetcollection}




\Section{Related Work}
\label{relatedwork}
\subsection{Image-Based Dietary Assessment}
Food is an important component of daily life. The type of foods and amount consumed can directly impact people's health. The recent success of modern deep learning techniques~\cite{resnet, densenet, alexnet} have greatly improved the performance of image-based diet assessment in recent years. 

\textbf{Food Classification.}
The most common food image recognition method is to apply state-of-the-art models~\cite{resnet, densenet, alexnet} to train a deep network that can recognize a variety of food items.
For example, authors in \cite{yanai2015food} use UEC-100~\cite{uec-100} and UEC-256~\cite{uec-256} food image datasets for testing, and ImageNet-ILSVRC~\cite{IMAGENET1000} for training. Their methods contain a combination of baseline feature extraction and neural network fine-tuning. An ensemble of deep networks are proposed in~\cite{foodnet-pandey2017} to improve the classification performance. A novel deep learning-based food image recognition algorithms is proposed in~\cite{deepfood-liu2016}, which is inspired by~\cite{googlenet,LeNet-5}.

\textbf{Food Portion Estimation.}
Automatic estimation of food portion size from an input food image is an open problem and there are many different methods to address it.
In~\cite{aizawa_2013}, food portion is divided into discrete serving sizes and food portion estimation is treated as a classification problem to determine the fixed serving size.
\cite{fang_2015} uses pre-defined 3D food models that are projected onto the scene to find the best fit with camera calibration.
In~\cite{murphy_2015}, food volume is estimated from the predicted depth map of the eating scene. The depth map is then converted to voxel representation which is used to estimated food volumes. An end-to-end approach for food energy estimation is proposed in~\cite{fang2019end}, where the concept of energy distribution map~\cite{icip2018} replaces the `depth map' in~\cite{murphy_2015} and the final food energy estimation is reported.

\subsection{Multi-task Learning} 
Multi-task learning~\cite{multi_task_cnn} (MTL)  has been applied to many computer vision problems that intended to impose knowledge sharing while solving multiple related tasks simultaneously. In the context of deep learning, MTL is typically done with either hard or soft parameter sharing of hidden layers. 

\textbf{Hard parameter sharing} is the most common method used in MTL where all tasks share the feature extraction layers while keeping task-specific output layers. 
In \cite{IBM}, the authors used MTL to improve the classification performance by clustering visually similar foods together. 
In~\cite{yanai_2017}, the authors applied MTL for food attribute prediction including food classes, ingredients, cooking instruction and food energy. However, sharing the feature map for cross domain tasks greatly impact the performance. In addition, the dataset used in~\cite{yanai_2017} for food energy is obtained by web crawler from a cooking website and cannot be verified for its accuracy. 

\textbf{Soft parameter sharing} is another approach in MTL where each task has its own model with its own parameters and the distance between the parameters of lower layers is then regularized in order to force the parameters to be similar. \cite{SPS_L2} proposed to use L2 distance for regularization and then \cite{SPS_TRACE} used the trace norm.


\Section{Dataset Collection}
\label{datasetcollection}
The performance of modern deep learning based methods greatly rely on the availability of good datasets, particularly datasets with correct annotation  for computer vision problems such as object recognition and detection. In this work, we aim to build a deep learning framework that can achieve the food classification and portion size estimation simultaneously. However, currently there is no available food related public dataset that contains both the groundtruth food categories and corresponding portion sizes. Therefore, we introduce an eating occasion image to food energy dataset that is collected from a nutrition study. The groundtruth portion size is provided by registered dietitians.

\subsection{Eating Occasion Image to Food Energy Dataset}
\begin{figure}[htbp]
\begin{center}
  \includegraphics[width=1.\linewidth]{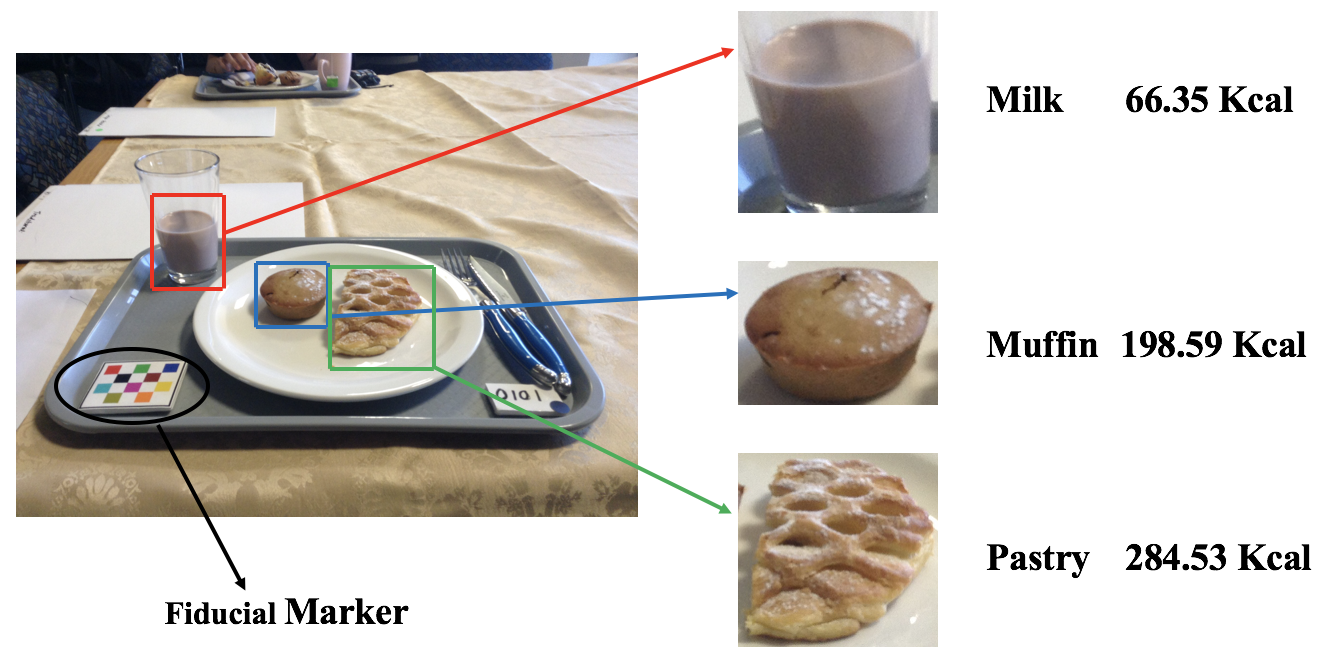}
  \caption{An example of eating occasion image: each food item is manually cropped containing corresponding groundtruth food category and portion size. A fiducial marker in left bottom is used to calibrate the color and size of the input image~\cite{FM}.}
  \label{fig:datasetexample}
\end{center}
\end{figure}

The dataset is collected as part of an image-assisted 24-hour dietary recall (24HR) study~\cite{boushey_2017new} conducted by registered dietitians. The study participants are healthy volunteers aged between 18 and 70 years old. A mobile app is used to capture images of the eating scenes for 3 meals (breakfast, lunch and dinner) over a 24-hour period. Foods are provided in buffet style in which pre-weighted foods and beverages in certain categories are served to the participants and they are asked to capture the eating scene images before they start to eat for each meal. The food energy is calculated and used as groundtruth. The dataset contains 96 eating occasion images and we manually crop each food item from each eating occasion as shown in figure~\ref{fig:datasetexample}. A total of 834 single food images belong to 21 categories are included in this dataset which contains both the category and portion size groundtruth.

\subsection{Balanced Data Augmentation}
Due to the lack of training data and unbalanced food images in each category, we implemented balanced data augmentation before the training step. Specifically, the operations performed are rotation (90 degrees, 270 degrees) and flip (x-axis, y-axis, both). We randomly implemented the operations based on the number of images for that category, i.e. we implemented less operations for the category which contains more images. In addition, we keep the groundtruth unchanged before and after the augmentation operations. Finally, we have 21 food categories, each contains around 100 images, so there are totally of 2,168 images. The groundtruth food energy of a single food item ranges from 0 kcal (diet coke) to 984 kcal and the mean is 164 kcal. We split the dataset into training and testing sets, with 1,744 and 424 images, respectively. 

\begin{figure*}[htbp]
\begin{center}
  \includegraphics[width=1.\linewidth]{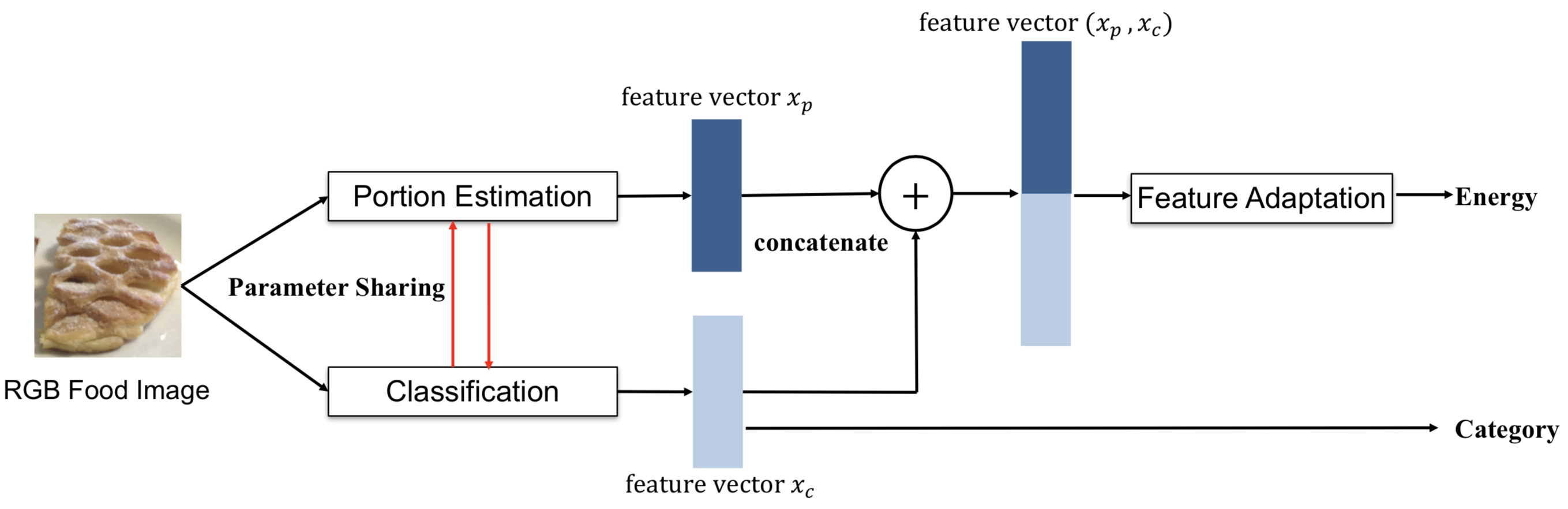}
  \caption{The architecture of our proposed model for image-based food classification and portion size estimation. L2-norm base soft parameter sharing is used to jointly train two tasks simultaneously. The feature vectors from each task are then concatenated together and we apply cross-domain feature adaptation to further improve the performance of food portion size estimation.}
  \label{fig:overall structure}
\end{center}
\end{figure*}

\Section{Our Method}
\label{ourmethod}
In this work, we propose an end-to-end framework for food classification and portion size estimation. The overall network structure is shown in figure~\ref{fig:overall structure}. 

\subsection{Multitask: Soft Parameter Sharing}
\label{SPS}
Multi-task Learning (MTL) is the most common method to simultaneously solve multiple tasks. Since we are performing two different tasks, i.e., classification and regression, using hard parameter sharing, where both tasks share the same feature map, is not suitable. More details will be described in Section~\ref{expresults}.  
Instead, we apply soft parameter sharing where each task has its own model with its own parameters. The distance between the parameters of lower layers of the two models is then regularized in order to force the parameters of the two models to be similar. The idea is that although the two tasks are different, they can be regarded as dependent tasks, i.e., the classification task can provide useful knowledge for optimizing regression task and at the same time the regression task can provide useful knowledge for learning classification task. This is based on the fact that it will be easier to know the food category if we know the food portion value at first and also it will be easier to get the food portion size if we know the food category \textit{a priori}. 

Given the input data $(\textbf{x}, y, z)$ where $\textbf{x}$ is the input image, $y$ and $z$ denoted the groundtruth for food category and portion size, respectively. We use cross-entropy loss $\mathcal{L}_c$ for classification and apply L1-norm loss $\mathcal{L}_r$ for portion size estimation. The two loss functions can be written as
\begin{equation}
    \label{two loss function}
    \begin{aligned}
    \mathcal{L}_c & = \sum_{i=1}^{n}-\hat{y}^{(i)}log[f_{c}^{(i)}(x)]\\
    \mathcal{L}_r & = |z - f_{r}(x)|
    \end{aligned}
\end{equation}
where $\hat{y}$ is the one hot label for food category and $n$ is the dimension of the feature vector. $f_c$ and $f_r$ denote the models for classification and regression, respectively. Note that for regression task, the vector dimension is 1. 

For parameter sharing, we use L2-norm to regularize the parameters of the two models. Let $p_c$ and $p_r$ denoted as the parameters of lower layers of classification model and regression model respectively, the loss function can be expressed as
\begin{equation}
    \label{ps loss function}
    \begin{aligned}
    \mathcal{L}_{ps} & = \sum_{i=1}^{m}(p_c^{(i)} - p_r^{(i)})^2
    \end{aligned}
\end{equation}
where $m$ is the size of parameters of two model. Note that since we apply the same network structure to the two tasks, we have the same number of parameters. 

Then, the overall loss function can be written as
\begin{equation}
    \mathcal{L}_{overall} = \mathcal{L}_c + \mathcal{L}_r + \mathcal{L}_{ps}
    \label{overall loss function}
\end{equation}

\subsection{Cross Domain Feature Adaptation}
\label{CDFA}
Different from classification task, it is difficult to map a RGB image to a numeric portion size value, e.g. if the input image is of size $224 \times 224 \times 3$, then direct approach would map $\mathcal{R}^{224\times224\times3} \to \mathcal{R}^{1\times1\times1}$ and it is difficult to learn such a mapping. Therefore, we concatenate the feature vector extracted using classification network as part of the feature vector extracted by the regression network. The feature vector for classification task can provide prior knowledge to assist the portion size estimation since it will be easier to estimate the food portion size if we already know the food category. We denote the features extracted from classification network as $\textbf{x}_c$ (of dimension $R^{512 \times 1}$) and the features extracted from the original portion estimation network as $\textbf{x}_p$ (of dimension $R^{512 \times 1}$). However, simply concatenating the features $(\mathbf{x}_{p}, \mathbf{x}_{c})$ (of dimension $R^{1024 \times 1}$) and applying fully-connected layers have fundamental issues. Features from the two domains have significant differences reflected by the mean and variance of the feature vectors. To adapt the features extracted from different domains and to remove imbalance in feature space for joint regression, we extensively studied the use of normalization techniques. 

In this work, we apply Batch Normalization (BN)~\cite{ioffe2015batch} and Layer Normalization (LN)~\cite{ba2016layer}. 
LN is defined as:
\begin{equation}
    \label{eq:LN}
    y_i = \gamma \hat{x}_i + \beta, \text{\ where \ } \hat{x}_i = \frac{x_i - \mu_L}{\sqrt{\sigma_L^2+\epsilon}} 
\end{equation}
where $\gamma$ and $\beta$ are learnable parameters, $\hat{x}_i$ is the normalized source domain sample for $x_i$ and $y_i$ is the mapped sample based on learned normalization. $\sigma_L$ and $\mu_L$ are defined as:
\begin{equation}
    \mu_L = \frac{1}{H} \sum_{i=1}^H x_i, \ \ \ \sigma_L^2 = \frac{1}{H} \sum_{i = 1}^H (x_i - \mu_L)^2
\label{eq:define_mu_std}    
\end{equation}
where $H$ denotes the number of hidden units in a layer. 

BN is defined as:
\begin{equation}
    \label{eq:BN}
    y_i = \gamma \hat{x}_i + \beta, \text{\ where \ } \hat{x}_i = \frac{x_i - \mu_B}{\sqrt{\sigma_B^2+\epsilon}} 
\end{equation}
Similarly $\gamma$ and $\beta$ are learnable parameters, $\hat{x}_i$ is the normalized source domain sample for $x_i$ and $y_i$ is the mapped sample. Let $B = \{x_1,...,x_m\}$ denote the mini-batch of input samples, $\sigma_B$ and $\mu_B$ are defined as:
\begin{equation}
    \mu_B = \frac{1}{m} \sum_{i=1}^m x_i, \ \ \ \sigma_B^2 = \frac{1}{m} \sum_{i=1}^m (x_i - \mu_B)^2
\label{eq:define_mu_std for BN}    
\end{equation}

\begin{table*}[htbp]
\begin{minipage}{1.0\linewidth}
\centering
\begin{tabular}{|c|c|c|c|c|c|c|c|c|c|}
\hline
\multicolumn{1}{|c|}{Method} & \multicolumn{1}{|c|}{Accuracy (\%)}& \multicolumn{1}{|c|}{MAE (kcal)}& \multicolumn{1}{|c|}{MAE-Correct (kcal)}& \multicolumn{1}{|c|}{MCCR}\\
\hline
Classification & 86.08 & - & - & -\\
Portion Estimation & - &62.27 & - & -\\
HPS & 50.23 &62.53 & - & -\\
SPS & 84.96 &63.51 & - & -\\
SPS+CDFA & 85.14 &66.64 & 61.10 & 0.7091\\
SPS+CDFA+BN & 86.32 &57.94 & 57.45 & 0.6577\\
SPS+CDFA+LN & 80.42 &62.94 & 54.83 & 0.6736\\
\textbf{SPS+CDFA+LN+BN} & \textbf{88.67} & \textbf{56.82} & \textbf{50.86} & \textbf{0.5667}\\
\hline
\end{tabular}
\end{minipage}
\caption{Experimental results for food classification and portion size estimation on food image dataset. The first two rows indicate the results by independently training two tasks. HPS and SPS denoted hard/soft parameter sharing multitask network respectively. CDFA corresponds to using cross domain feature adaptation. LN and BN refer to layer normalization and batch normalization (Best results marked in bold).}
\label{table:allresult}
\end{table*}

\Section{Experimental Results}
In this part, we evaluate the performance of our proposed method using the dataset introduced in section~\ref{datasetcollection}. For portion estimation,  we use Mean Absolute Error (MAE), defined as 
\begin{equation}
\text{MAE} = \frac{1}{N}\sum^{N}_{i=1} |\tilde{w}_i - \bar{w}_i|
\label{eq:mae}
\end{equation}
where $\tilde{w}_i$ is the estimated portion value of the $i$-th image, $\bar{w}_i$ is the groundtruth portion size of the $i$-th image and $N$ is the number of testing images. We use accuracy to evaluate classification performance. However, since have a multi-task for both classification and regression, we need a better metric that can balance the performance of MAE and classification accuracy. We propose a new metric called MAE to Correctly Classified Ratio (MCCR):
\begin{equation}
\text{MCCR} = \frac{C\sum_{i\in I} |\tilde{w}_i - \bar{w}_i|}{||I||^2}
\end{equation}
where $I$ denote the correctly classified image. $C$ is a constant, in this experiment, we use $C=1$. Note that we only calculate the mean absolute portion size error for correctly classified food in this new metric since if the classification result is wrong then it is meaningless to give an estimated portion size. The multi-task network has better performance when the metric has a smaller value.

\subsection{Implementation Detail}
Our implementation is based on Pytorch~\cite{pytorch}. We use standard 18-layer ResNet and the ResNet implementation follows the setting suggested in~\cite{resnet}. We train the network for 100 epochs using Adam optimizer. The learning rate is set to 0.1 and reduces to 1/10 of the previous learning rate after 30, 60, 90 and epochs. The weight decay is set to 0.0001 and the batch size is 32.

\subsection{Evaluation of Our Proposed Method}
\label{expresults}
Results are shown in Table~\ref{table:allresult}. Compared to the two baseline methods that separately train two networks for portion estimation and classification, our method improves both the classification accuracy and the mean absolute error for estimated portion size. In addition, we show that directly using the concatenating features $(\mathbf{x}_{p}, \mathbf{x}_{c})$ causes the performance degradation in MAE since the features from two domain have significant differences reflected by the mean and variance of the two feature vectors from two tasks. We also compared the results using three normalization methods, BN, LN and LN+BN. As shown in Table~\ref{table:allresult}, by using LN+BN, we are able to achieve the best classification accuracy and MAE. For correctly classified food, the MAE is only 50.86 Kcal.

\subsection{Portion Estimation Comparison to Human Estimates}
We also want to compare our portion estimation results with the participants' estimation from the same nutrition study. At each recorded meal, participants estimated the portion size of the meal they consumed in a structured interview while viewing the captured images. The error percentage is defined as $EP = \frac{\sum_{i=0}^N|w_i-\hat{w}_i|}{\sum_{i=0}^N\hat{w}_i}\times 100\%$ where $w_i$ is the estimated portion size and $\hat{w}_i$ is the groundtruth portion size. The EP of participant estimates is 45.43\%, we compared this human results to our best result by using cross-domain feature adaptation together with LN+BN normalization, which is 16.83\%. The comparison shows that our proposed method outperforms human estimation, indicating that estimating portion size accurately from a food image is a very challenging task for human.

\Section{Conclusion}
In this work, we proposed a multi-task framework for food classification and food portion size estimation by using L2-norm based soft parameter sharing. We also investigated cross-domain feature adaptation together with different normalization techniques to further reduce portion estimation error. Our method is evaluated on a real life eating occasion food image dataset with groundtruth category and portion size provided by registered dietitians. Our best result achieved 88.67\% classification accuracy, with the mean absolute errors of 56.82 Kcal for all food and 50.86 Kcal for correctly classified food for portion size estimation, surpassing the baseline results which are 86.08\% and 62.27 Kcal respectively. In addition, we compared our portion estimation results with human estimates, showing an impressive 28.57\% reduction in error percentage. 
\label{conclusion}

\bibliographystyle{latex8}
\bibliography{latex8}

\end{document}